# Accelerating gradient-based topology optimization design with dual-model neural networks

Chao Qian,  Wenjing Ye*

Department of Mechanical and Aerospace Engineering, The Hong Kong University of Science and Technology, Clear Water Bay, Kowloon, Hong Kong

*Corresponding author, mewye@ust.hk

**Abstract**: Topology optimization (TO) is a common technique used in free-form designs. However, conventional TO-based design approaches suffer from high computational cost due to the need for repetitive forward calculations and/or sensitivity analysis, which are typically done using high-dimensional simulations such as Finite Element Analysis (FEA). In this work, neural networks are used as efficient surrogate models for forward and sensitivity calculations in order to greatly accelerate the design process of topology optimization. To improve the accuracy of sensitivity analyses, dual-model neural networks that are trained with both forward and sensitivity data are constructed and are integrated into the Solid Isotropic Material with Penalization (SIMP) method to replace FEA. The performance of the accelerated SIMP method is demonstrated on two benchmark design problems namely minimum compliance design and metamaterial design. The efficiency gained in the problem with size of 64x64 is 137 times in forward calculation and 74 times in sensitivity analysis. In addition, effective data generation methods suitable for TO designs are investigated and developed, which lead to a great saving in training time. In both benchmark design problems, a design accuracy of 95% can be achieved with only around 2000 training data.

## 1. Introduction:

Topology optimization is a mathematical technique commonly used in free-form designs. Since its invention (Bendsøe and Kikuchi 1988), various TO-based design approaches have been developed (Jakiela et al. 2000, Wang, M. Y. et al. 2003, Juan et al. 2008, Schevenels et al. 2011, Guo et al. 2014, Zhang, W. et al. 2017, Zhang, X. et al. 2019, Zhao et al. 2020) and applied to design a wide range of structures and products such as automobile and aircraft parts/components (Cavazzuti et al. 2011, Zhu et al. 2016). The advent in additive manufacturing technologies has further broadened the application scope of TO. Advanced materials such as phononic materials (Sigmund and Jensen 2003), various metamaterials (Diaz and Sigmund 2010, Matsumoto et al. 2011, Rong and Ye 2019) and artificial bone scaffolds and orthopaedic implants (Wang, X. et al. 2016), have been successfully designed using TO methods.

However, almost all TO methods are cursed with the exorbitant computational cost. The major bottleneck in the TO process is the repetitive evaluation of the objective function, constraints and/or sensitivities as the structure evolves. These evaluations, which require the modeling of the underlying physical problem, are often conducted using high-dimensional numerical simulations such as finite element analysis. For large-scale design problems, these calculations can be very costly and typically consume more than 80% of the computational resources. The situation becomes worse when a series of structures/materials are to be designed, for example, the design of compliant structures with different loading conditions or the design of functionally graded materials in which a set of microstructures that have gradually changing effective properties are to be designed. The high computational cost severely limits the design resolution and also the application scope of TO methods.

The need for efficient methods to speed up the design process has been well recognized in the field (Aage and Lazarov 2013, Groen et al. 2017). Efficient solution schemes and new methods, for example, multiscale finite element methods (Liu, H. et al. 2018), multi-resolution TO(Nguyen et al. 2012, Groen et al. 2017, Liu, C. et al. 2018) and others (Guo et al. 2016, Li et al. 2018), have been developed to reduce the computational cost. Strategies utilizing computer hardware such as parallel computing techniques (Aage et al. 2017, Martínez-Frutos et al. 2017) have also been proposed. A brief review of various methods and strategies can be found in (Liu, H. et al. 2018). Typically these solution schemes/methods result in one order of magnitude improvement in computational efficiency for 3D problems.

In recent years, machine learning techniques, particularly deep learning methods, have achieved remarkable success in a wide range of applications. Some of these techniques have also been applied to solving topology optimization problems. One group of approaches use various machine learning techniques to speed up the evaluation of the objective function. For instance, in topology optimization of structural acoustics artificial neural network is used to approximate the objective function (Ranjbar and Marburg 2013) and their artificial neural network's forward prediction performance is investigated with different parameter settings (Ranjbar and Saffar 2016). In the design of composite materials for high toughness and strength, Gu et al. (2018) used a one-layer neural network (NN) to predict the toughness and strength of a bi-phase composite material represented by a 16x16 grid with each element in the grid being one of the two phases of the material. Owing to the simplicity of the neural network, the weights of the network characterize the contribution of each element on the material toughness/strength. Based on this information, composite materials with the highest toughness or strength were found. In a work by Raponi et al. (2019), a kriging surrogate model was constructed and used to predict the compliance of structures in the design of cantilever beams with minimum compliance under the moving morphable components framework. Neural networks have also recently been incorporated in the multiscale topology optimization framework in which a one-layer NN was used to predict effective material properties of a microstructured metamaterial (White et al. 2019). In most of the work, the design space is relatively small and simple surrogate models such as one-layer NN are sufficient to provide accurate predictions and guide the design since sensitivity information can be readily obtained. When the size of the topology optimization problem is large and the underlying structure becomes complex, the corresponding neural network will have to be much deeper in order to provide good predictions. Consequently the sensitivity information will become difficult to obtain accurately. Normally, back propagation is employed to obtain derivatives of network outputs with respect to its inputs. As shown in Section 3 and also pointed out by White et al. (2019), such an approach does not provide accurate sensitivity information needed in the design. Thus an accurate and efficient method must be developed to evaluate sensitivity information. Another issue is regarding the efficiency of the method. Although neural networks can typically provide several orders of magnitude improvement in computational efficiency as compared with high-dimensional simulations, for example, the finite element analysis, the training of these neural networks requires significant computational resources because (1) the need of a large number of training data and (2) high-dimensional simulations are often used to generate groundtruth values. Although conducted off-line, it nevertheless adversely affects the efficiency of design methods. This issue has been largely overlooked in most of the existing work. To truly improve the efficiency, effective data generation methods must be developed.

    Another group of methods are to use various neural networks to directly produce design solutions. For example, convolutional neural networks (CNN) were used to produce the final optimized structure directly from an intermediate design, thus shortening the optimization process (Sosnovik and Oseledets 2017, Banga et al. 2018). Under the moving morphable component-based framework, Lei et al. (2019) used principal component analysis and support vector machine/K-nearest-neighbors algorithms to map loading conditions directly to design parameters for the design of compliance structures. Recently with the advent in image generative neural networks such as generative adversarial neural network (GAN) and variational autoencoder (VAE), generative machine learning methods are gaining popularity in topology optimization studies. Yu et al. (2018) used a CNN-based encoder and decoder to directly map the initial structure, boundary conditions and other constraints to a low-resolution near-optimal compliance structure. A conditional GAN was then used to map the low-resolution design to the high-resolution final design. A common issue in these approaches is the high computational cost associated with the generation of a large corpus of the training data, which are a set of optimal structures corresponding to different loading conditions/boundary conditions obtained from conventional optimization methods. In a work by Zhang et al. (2018), a VAE model was used to perform layout design. The VAE model serves two functions. One is to generate design candidates and the other is to correlate the layout with the response. In a later paper (Tan et al. 2019), the two functions were separated to

improve the performance. A GAN was used to generate design candidates and a CNN was used to map the design to its mechanical response/functionality, for example, compliance. A design neural network was then constructed by connecting the generator of the GAN with the CNN to produce optimal designs directly. The advantages of these two approaches are that (1) the generation of training data requires only forward calculations and (2) shape constraints can be easily imposed in the topology optimization design because generative models can learn shapes/features in the training data and generate new images with the same shape/features. It should be noted that the optimal design obtained by this approach is confined within the design space generated by the GAN. Thus to obtain the global optimal design, the design space generated by the GAN must contain this optimal structure.

This work concerns with the first type of approaches. In particular, we develop dual-model neural networks that provide accurate evaluations of objective functions and sensitivities. These neural networks are then incorporated into a gradient-based TO framework to replace high-dimensional simulations. In addition, effective structure generation and augmentation methods have been developed to reduce the training cost. The performance of the developed accelerated TO method is demonstrated on two benchmark design problems: the design of a series of structures with minimal compliance subjected to different loading conditions and the design of a set of metamaterials with prescribed negative Poisson's ratios and volume fractions.

In the next section, the SIMP method is briefly reviewed. Two benchmark design problems are introduced and the corresponding TO formulations are described. Section 3 describes dual-model neural networks developed for forward calculation and sensitivity analysis and data generation methods. Design solutions obtained from the accelerated SIMP method are presented in Section 4. Performance comparison between the accelerated method and the FEM-based SIMP method is also presented. Section 5 provides the summary of the work.

## 2. Problem description and TO formulation:

In the SIMP method, the design domain is discretized into many small elements. Each element is assigned a continuous density value between 0 and 1, where 0 and 1 represent two different phases respectively. For single-phase structure design, the local Young's modulus of each element is defined as follows:

$$E(\rho_e) = \rho_e^P (E_0 - E_{min}) + E_{min}, \quad p > 1 \qquad (1)$$

where $\rho_e$ is the element density, p is the penalization parameter, usually $p = 3$, and $E_0$ is the Young's modulus of the solid material. $E_{min}$ is a small value assigned to void regions to avoid the singularity of the global stiffness matrix.

A general topology optimization formulation for density based structure design is listed as follows:

$$\underset{\rho_e}{\text{minmize:}} f(\boldsymbol{u}(\rho_e), \rho_e)$$

$$\text{Subject to: } K(\rho_e)U = F$$

$$\text{other constraints} \quad (2)$$

$$0 \leq \rho_e \leq 1, \quad e = 1, \ldots, N$$

where $f(\boldsymbol{u}, \rho_e)$ is the objective function, $\boldsymbol{u}$ is a state field that satisfies a linear or nonlinear state equation, $\rho_e$, is the density distribution, which is also the design variables, $KU = F$ is the equilibrium equation that needs to be satisfied and $N$ is the total number of element. This problem is solved using an optimization scheme. In gradient-based optimization algorithms, high-dimensional simulations are often used to evaluate the objective function, $f(\boldsymbol{u}, \rho_e)$, and perform the sensitivity analysis, $\frac{\partial f}{\partial \rho_e}$. The sensitivity filtering technique proposed by Bendsøe and Sigmund (2003) is used to reduce mesh dependency, and the filter radius is 4 pixels in this work. The density of each element is then updated based on filtered sensitivities $\widetilde{\frac{\partial f}{\partial \rho_e}}$ and the optimization scheme, for example, the grdient descent method used in this work. This process is repeated until a convergence is achieved. To obtain a binary design with good manufacturability, density projection scheme proposed by Wang et al. (Wang, F. et al. 2011) can be used to reduce intermediate densities. In this work, the sharpness parameter *η* is initially set as 1 and enlarges gradually while the threshold parameter *β*=0.5.

## 2.1 Minimal compliance design

Structure design with minimal compliance is a classical design problem used to demonstrate the performance of many TO methods (Suzuki and Kikuchi 1991, Sigmund 2001, Luo et al. 2009). The design objective is to design a structure that is as stiff as possible with a given amount of material and subjected to given loading and boundary conditions. In this work, a serial of cantilever structures with different loading conditions and volume fractions are to be designed. The range of the volume fraction is set to be [0.3, 0.7]. The applied load is a vertical distributed load with a fixed magnitude. But the location of the force varies along the right edge of the design domain. The left boundary is fixed and all other sides are free.

The TO formulation of this problem is:

$$\underset{\rho}{\text{minimize:}} C(\boldsymbol{u}, \rho_e) = \int_\Omega C_{ijkl}\varepsilon_{ij}\varepsilon_{kl}d\Omega = \int_\Omega E_e \boldsymbol{u}_e^T \boldsymbol{K}_e \boldsymbol{u}_e d\Omega \quad (3)$$

$$\text{subject to: } G_0(\rho) = \left(\frac{1}{\Omega}\int_\Omega \rho(x)d\Omega - V_0\right)^2 = 0 \quad (4)$$

where $C(\boldsymbol{u}_e, \rho_e)$ is the strain energy, $C_{ijkl}$ is the stiffness tensor, $\varepsilon_{ij}$ is the strain tensor, $\boldsymbol{u}_e$ is the displacement vector containing the displacement of each element, $\boldsymbol{K}_e$ is the stiffness matrix, $E_e$ is the local Young's modulus, and $V_0$ is the target volume fraction.

This constrained optimization problem can be converted into an unconstrained problem with its objective function defined as follows:

$$\underset{\rho}{\text{minimize:}} F = C(\boldsymbol{u}_e, \rho_e) + \lambda G_0(\rho) = \int_\Omega E_e \boldsymbol{u}_e^T \boldsymbol{K}_e \boldsymbol{u}_e d\Omega + \lambda (\frac{1}{N}\sum_i \rho_i - V_0)^2 \quad (5)$$

where $\rho_i$ is the density of the $i$-th element and $\lambda$ is a Langrange multiplier. The gradient of the objective function with respect to the density distribution can be analytically expressed as:

$$\frac{DF(\boldsymbol{u}_e, \rho_e)}{D\rho_e} = -\frac{\partial E_e}{\partial \rho_e}\boldsymbol{u}_e^T \boldsymbol{K}_e \boldsymbol{u}_e + \lambda \frac{2}{N}(\frac{1}{N}\sum_i \rho_i - V_0) \quad (6)$$

## 2.2 Design of metamaterials with negative Poisson's ratio

Metamaterials are artificial materials with unique properties. Through the design of the microstructure of their building blocks, novel properties can be achieved. In the second benchmark example, a set of 2D metamaterials with desired Poisson's ratios in the range of [-0.6, 0] and given volume fractions, $V_0$, in the range of [0.3, 0.65] will be designed. This design problem can be formulated as an TO problem described in Eqn. (7)

$$\underset{\rho}{\text{minimize }} F_1 = (v_{12} - v_0)^2$$

$$\text{subject to } G_0(\rho) = \left(\frac{1}{\Omega}\int_\Omega \rho(x)d\Omega - V_0\right)^2 = 0 \quad (7)$$

where $v_0$ is the desired Poisson's ratio and $v_{12}$ is the effective Poisson's ratio of the metamaterial defined as the ratio of two stiffness components, $C^H_{1111}$ and $C^H_{1122}$:

$$v_{12} = \frac{C^H_{1122}}{C^H_{1111}} \quad (8)$$

$C^H_{ijkl}(i,j,k,l = 1,2)$ denotes the effective stiffness tensor that relates macroscopic stress, $\sigma^M_{ij}$, and macroscopic strain, $\varepsilon^M_{kl}$, where the superscript M denotes macroscopic quantities. To ensure certain rigidity of the material, the effective stiffness components in two directions, $C^H_{1111}$ and $C^H_{2222}$, are required to be as high as possible. In addition, isotropic material is desired. Combining all the requirements together, the objective function of this design problem is formulated as

$$\underset{\rho}{\text{minimize }} F = \lambda_1(\frac{C^H_{1122}}{C^H_{1111}} - v_0)^2 + \lambda_2(\frac{1}{N}\sum_i \rho_i - V_0)^2 + \lambda_3 \frac{(C^H_{1111} - C^H_{2222})^2}{E_0^2} - \lambda_4 \frac{C^H_{1111} + C^H_{2222}}{E_0} \quad (9)$$

where $\lambda_i (i = 1, 2, 3, 4)$ are the weighting parameters. Initially they are set as 1.0, 1.0, 2.0 and 0.1 respectively. Since the main objective is to obtain structures with the desired Poisson's ratio and volume fraction, $\lambda_1$ and $\lambda_2$ gradually enlarge every 25 iterations to reinforce these requirements if the target Poisson's ratio and volume fraction have not been achieved.

The effective stiffness tensor can be calculated using the homogenization method documented in (Sigmund 1994, Xia and Breitkopf 2015, Kouznetsova et al. 2001). Based on the mutual energy conservation, the stiffness tensor is given by

$$C_{ijkl}^H = C_{pqrs}^H \varepsilon_{pq}^{M(ij)} \varepsilon_{rs}^{M(kl)} V = \int_V C_{pqrs} \varepsilon_{pq}^{m(ij)} \varepsilon_{rs}^{m(kl)} dV = \int_V E_e (\boldsymbol{u}_e^{m(ij)})^T \boldsymbol{K}_e \boldsymbol{u}_e^{m(kl)} dV \quad (10)$$

where $C_{pqrs}$ is the stiffness of the solid material, $\varepsilon_{pq}^{m(ij)} (p, q = 1,2)$ is the microscale strain and $u_e^{m(ij)}$ is the element displacement corresponding to a given macroscale testing strain $\varepsilon^{M(ij)}$.

The gradient of the objective function with respect to the density of each element can be calculated using chain rule:

$$\frac{DF(C_{ijkl}^H, \rho_e)}{D\rho_e} = \frac{\partial F}{\partial C_{ijkl}^H} \frac{\partial C_{ijkl}^H}{\partial \rho_e} + \frac{\partial F}{\partial \rho_e} \quad (ij, kl = 11,22) \quad (11)$$

where $\frac{\partial C_{ijkl}^H}{\partial \rho_e}$ can be obtained using adjoint analysis (Bendsøe and Sigmund 2003),

$$\frac{\partial C_{ijkl}^H}{\partial \rho_e} = \frac{\partial E_e}{\partial \rho_e} (\boldsymbol{u}_e^{m(ij)})^T \boldsymbol{K}_e \boldsymbol{u}_e^{m(kl)}, \quad (12)$$

and $\frac{\partial E_e}{\partial \rho_e}$ can be calculated from Eqn. (1).

## 3. Dual-model neural networks for forward calculation and sensitivity analysis

Artificial neural networks create analytical mappings between the input and the output and are used to approximate complex relationships between two sets of data. In recent years, neural networks have been used increasingly in physical systems to serve as efficient surrogate models. Due to its analytical nature, the sensitivities of the output with respect to the input can be obtained by back-propagating the neural network. However, the accuracy of these sensitivities is limited by the accuracy of the neural network in its forward prediction. As an example, a 4-layer fully connected neural network shown in Figure 1 was constructed and trained to predict the effective stiffness tensor of a 2D symmetric microstructure. A quarter of the 2D microstructure was embedded in a square domain and casted into a 32x32 image file. Each pixel value represents the density of the corresponding material element. This image was then reshaped into a 1D array and input into the network. The prediction accuracy of the stiffness component $C^H_{1111}$ is very high as shown in Figure 2 where stiffness components of various structures predicted from the network are plotted against ground-truth values. The average accuracy reaches to 99.72%. However the accuracy of the sensitivity of $C^H_{1111}$ with respect to each pixel value obtained from back-propagation is unsatisfactory as indicated in Figure 2. Most of data points are far away from ground-truth values as indicated by the red line.

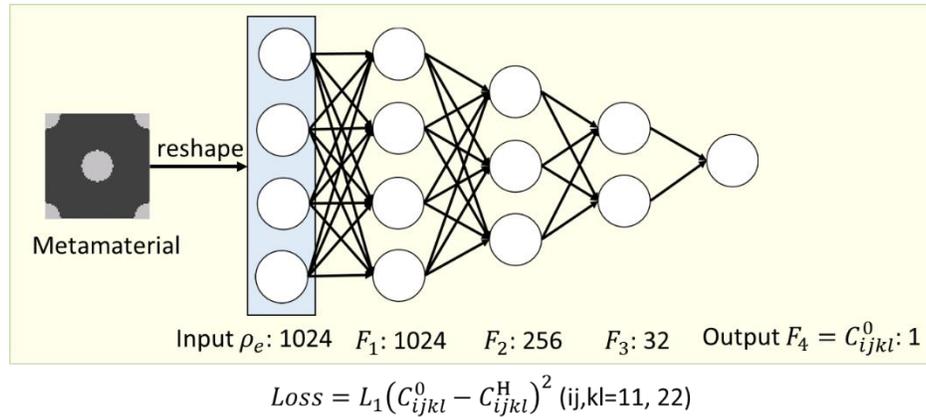

Figure 1: The architecture of the fully connected neural network used to predict stiffness tensors of 2D microstructures.

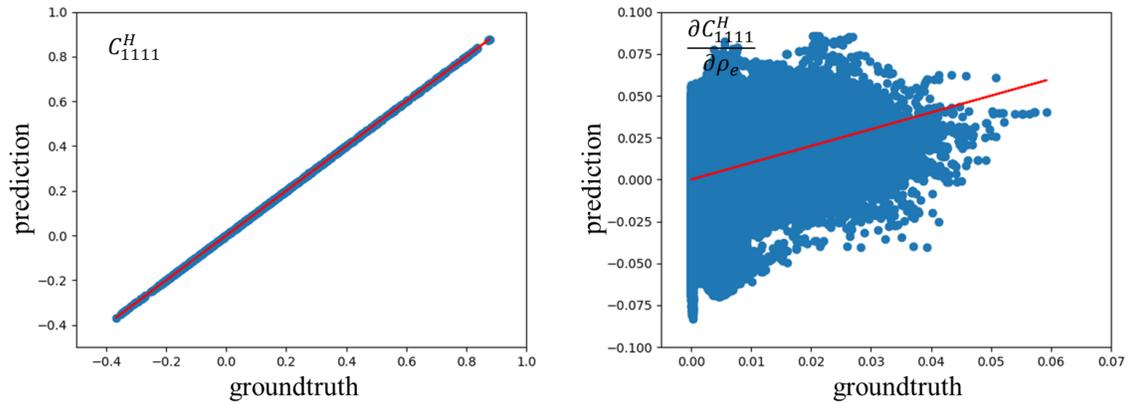

Figure 2: $C^H_{1111}$ obtained from the neural network shown in Figure 1 (left) and the corresponding sensitivity with respect to pixel value, $\frac{\partial C^H_{1111}}{\partial \rho_e}$, obtained from back-propagation(right) versus their groundtruth values. The red line in both plots has a slope of one, indicating a perfect match between predicted values and ground-truth values.

To improve the prediction accuracy of sensitivity, a dual-model neural network, which is trained to produce accurate results for both forward prediction and sensitivity calculation, is employed in this work. This is done by adding the loss corresponding sensitivity into the original loss function, that is, Loss = the square error of the forward prediction+ the square error of the sensitivity prediction. It should be pointed out that a similar idea was employed in the work of Xu (Xu et al. 2003) and White et al. (White et al. 2019) to produce more accurate results. To facilitate the training of the network, an adjoint neural model is used to compute sensitivities. The adjoint network is essentially the inverse net of the forward model but with a linear activity function and different weights. The value of each neuron in the adjoint network is the sensitivity of the output with respect to the value of the corresponding neuron in the forward network. The weights are the local derivatives of neurons of two neighboring layers in the forward network.

As an illustration, a simple 3-layer fully connected network and its corresponding adjoint network are illustrated in Figure 3. The value of each neuron in the forward network is denoted as $x_i^j$, where the superscript $j$ indicates the layer number and the subscript $i$ is the index of neurons in each layer. Denoting $g^j$ as the activation function between the $j$-th layer and the $(j+1)$-th layer, and $w_{i,k}^j$ as the weight connecting the $i$-th neuron in the $j$-th layer and the $k$-th neuron in the $(j+1)$-th layer, the value of each neuron in the forward model can be computed as

$$x_i^{j+1} = g^j(\sum_{k=1}^{N^j} x_k^j w_{k,i}^j) \qquad (13)$$

where $N^j$ is the total number of neurons in the $j$-th layer.

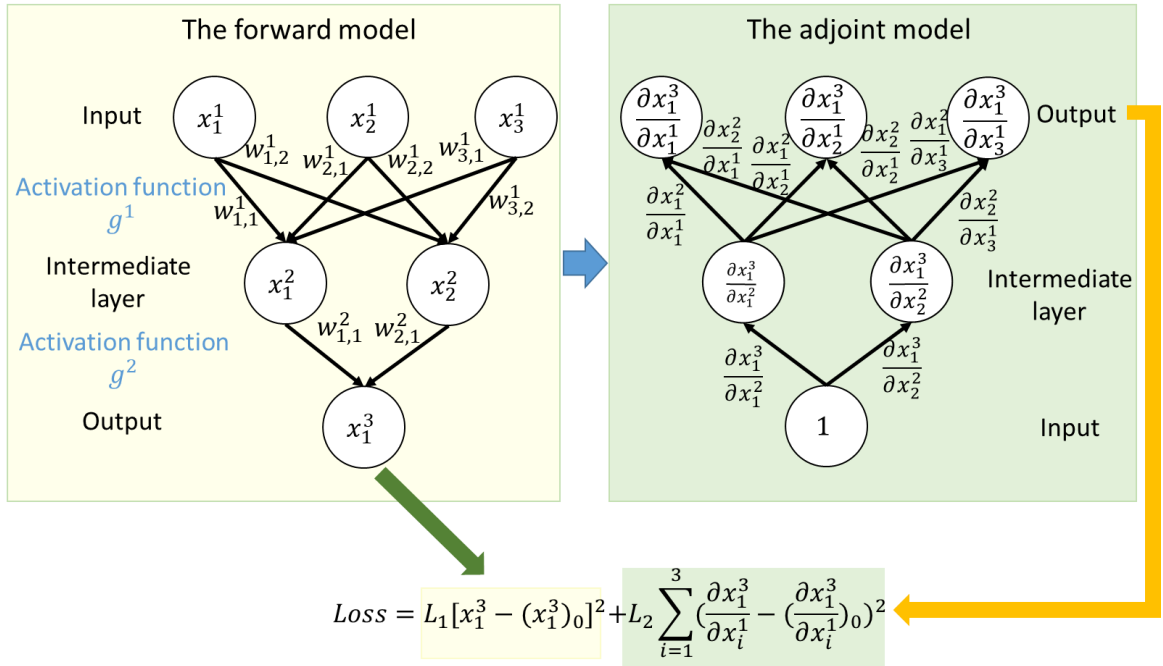

Figure 3: Schematics of a 3-layer dual-model neural network.

The architecture of the adjoint model is the reverse of the forward model with one input neuron and three output neurons. Each neuron in the adjoint model represents the derivative of the output of the forward model, that is, $x_1^3$, with respect to the corresponding neuron in the forward model. For example, the value of the first output neuron in the adjoint model is $\frac{\partial x_1^3}{\partial x_1^1}$ and the value of the second neuron in the second layer is $\frac{\partial x_1^3}{\partial x_2^2}$. The input neuron represents $\frac{dx_1^3}{dx_1^3}$, and hence its value is 1. These neuron values can be calculated sequentially from the input neuron to output neurons following the chain rule. For example:

$$\frac{\partial x_1^3}{\partial x_j^1} = \frac{\partial x_1^3}{\partial x_1^2} * \frac{\partial x_1^2}{\partial x_j^1} + \frac{\partial x_1^3}{\partial x_2^2} * \frac{\partial x_2^2}{\partial x_j^1} \qquad j = 1,2,3 \tag{14}$$

where $\frac{\partial x_1^3}{\partial x_1^2}$ and $\frac{\partial x_1^3}{\partial x_2^2}$ are neurons in the previous layer and $\frac{\partial x_1^2}{\partial x_j^1}$ and $\frac{\partial x_2^2}{\partial x_j^1}$ are the local derivatives of neurons in the forward model, which can be obtained analytically. Hence the weights of the adjoint model are the local derivatives of neurons in the forward model and the activation function is the linear function.

To train the dual-model neural network for both forward prediction and sensitivity analysis, the loss function consists of two parts as expressed in Eqn. (15),

$$Loss = L_1[x_1^3 - (x_1^3)_0]^2 + L_2 \sum_{i=1}^{3} (\frac{\partial x_1^3}{\partial x_i^1} - (\frac{\partial x_1^3}{\partial x_i^1})_0)^2 \,, \tag{15}$$

where $x_1^3$ and $\frac{\partial x_1^3}{\partial x_i^1}$ are the predictions given by the dual-model neural network, $(x_1^3)_0$ and $(\frac{\partial x_1^3}{\partial x_i^1})_0$ are the groundtruth data. $L_1 = 1$ and $L_2 = 1e6$ are the weight coefficients, which can also be adjusted to balance the accuracy of the forward prediction and the sensitivity calculation.

For the example shown in Figure 2, a dual-model neural network was constructed to predict the effective stiffness tensor and its sensitivities of 2D microstructures. The predicted $C_{1111}^H$ and $\frac{\partial C_{1111}^H}{\partial \rho_e}$ of 1000 microstructures obtained from the dual-model network are plotted against the ground-truth values in Figure 4. Compared to Figure 2, the accuracy of sensitivity prediction $\frac{\partial C_{1111}^H}{\partial \rho_e}$ is much improved.

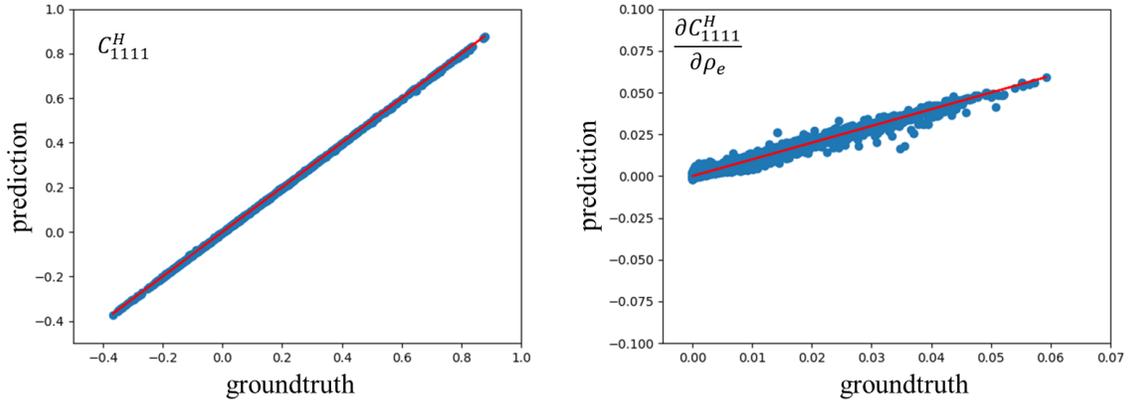

Figure 4: $C^H_{1111}$ (left) and the corresponding sensitivity with respect to pixel value, $\frac{\partial C^H_{1111}}{\partial \rho_e}$ (right) obtained from the dual-mode neural network versus their groundtruth values.

## 3.1 Dual-model convolutional neural network for the minimal compliance design

To design a cantilever structure with minimal compliance subjected to a vertical load with its location varied along the right edge of the design domain, a dual-model NN is constructed to compute the compliance of the structure using a forward model and its derivatives with respect to the density of each element using an adjoint model. The inputs to the forward model should be the structure and its loading condition. To handle such inputs, a convolutional neural network (CNN) shown in Figure 5 is constructed, which consists of two convolutional layers with kernel size 5 by 5 and 6 by 6 correspondingly and a 7-layer fully connected block. The linear activation function is used in the final layer while the activation function tanh is used in all the other layers. The density distribution of the structure and the force distribution are casted as a 2-channel image and is inputted to the CNN. The output is the compliance of the structure. The adjoint model is constructed following the method described above and is shown in Figure 5.

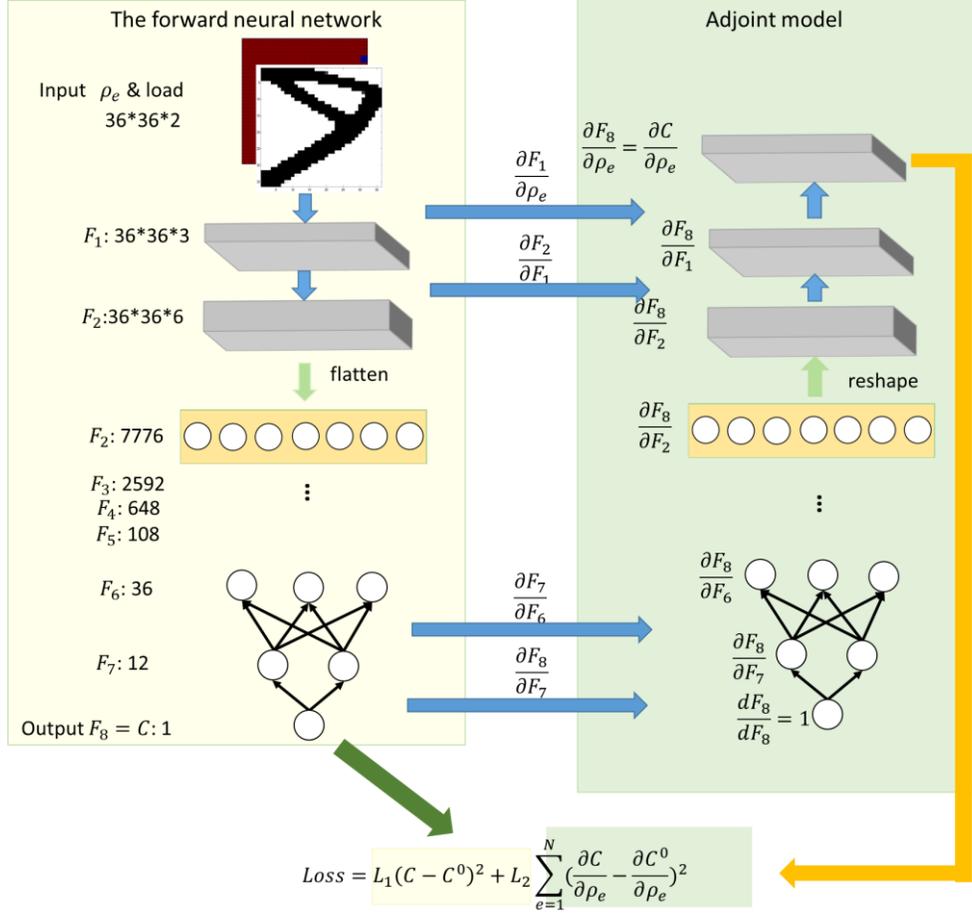

Figure 5: The dual-model convolutional neural network for forward prediction and sensitivity analysis

The loss function to train this dual-model CNN is defined as:

$$Loss = L_1(C - C^0)^2 + L_2 \sum_{e=1}^{N} \left(\frac{\partial C}{\partial \rho_e} - \frac{\partial C^0}{\partial \rho_e}\right)^2, \qquad (16)$$

where $L_1$ and $L_2$ are the two weighting parameters that can be adjusted according to the accuracy demands of the two parts, $C^0$ and $\frac{\partial C^0}{\partial \rho_e}$ $(ij, \ kl = 11, 22)$ are the groundtruth values of the structural compliance and its sensitivities calculated from the finite element analysis, $C$ and $\frac{\partial C}{\partial \rho_e}$ are the predictions by the neural network.

## 3.2 Dual-model neural network for the design of metamaterials with negative Poisson's ratios

For the design of metamaterials with negative Poisson's ratios, a dual-model fully connected network is constructed to predict the effective stiffness component of the material and its sensitivities with respect to the density of each element. The forward model is the same as the one shown in Figure 3 with its inputs being the density of each element in the design domain and the output being the effective stiffness component, $C_{ijkl}^H$. The activation function tanh is used in the first three layers, and the linear activation fucntion is used in the last layer. The corresponding adjoint model, which outputs the sensitivities of the stiffness component with respect to each density, $\frac{\partial C_{ijkl}^H}{\partial \rho_e}$, is also shown in Figure 6.

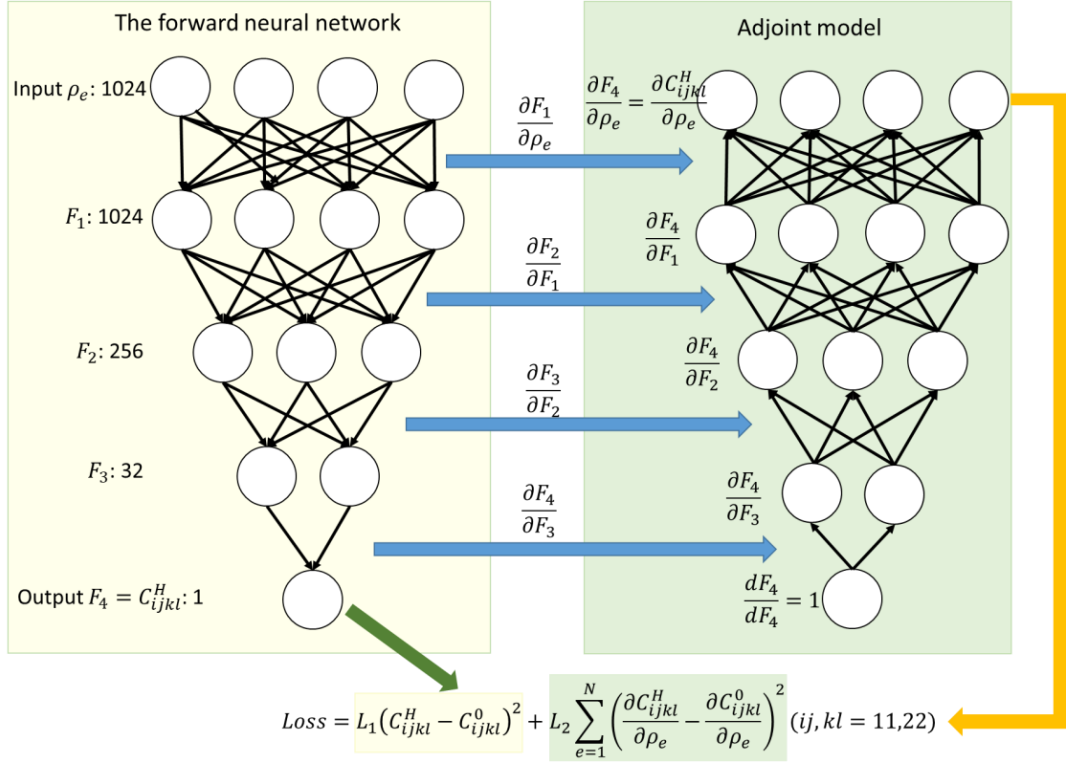

Figure 6: The dual-model neural network for forward prediction and sensitivity analysis

The loss function to train this dual-model neural network is:

$$Loss = L_1(C_{ijkl}^H - C_{ijkl}^0)^2 + L_2 \sum_{e=1}^{N} \left(\frac{\partial C_{ijkl}^H}{\partial \rho_e} - \frac{\partial C_{ijkl}^0}{\partial \rho_e}\right)^2, \quad ij, kl = 11, 22) \qquad (17)$$

where $L_1 = 1$ and $L_2 = 1e6$ are weighting parameters, $C_{ijkl}^0$ and $\frac{\partial C_{ijkl}^0}{\partial \rho_e}$ $(ij, \ kl = 11, 22)$, are the ground-truth values of the effective stiffness tensor and the sensitivities calculated from the FEA while $C_{ijkl}^H$ and $\frac{\partial C_{ijkl}^H}{\partial \rho_e}$ $(ij, \ kl = 11, 22)$, are the predicted values. Individual neural network is constructed for each component of the stiffness tensor $C_{ijkl}^H$.

## 3.3 Data generation

Unlike those typical application areas of machine learning in which abundant data are available, the training data required in the supervising learning of physical problems has to be obtained either from high-dimensional simulations or experimental measurements. Both approaches require significant time and efforts. Therefore efficient methods for data generation need to be developed. In this work, instead of generating random structures to form the dataset, the conventional topology optimization method is used to conduct the design for a few selected design cases. Structures generated during the design process are saved and form the dataset. The rationale of this approach is illustrated in Figure 7 in which design solutions corresponding two cases with the same loading condition but different volume fractions are shown. The high similarity between the two designs, particularly the topology, indicates that the proposed approach should be able to generate a dataset that is much smaller but as effective as the dataset consisting of random structures. In fact, it is not necessary to complete the entire design process and include all structures in the dataset. In most TO designs, the topology of the structure only evolves rapidly during the early design stage and maintains almost the same during the later stage of the design. Thus including structures generated from the early design stage is sufficient to form a good dataset.

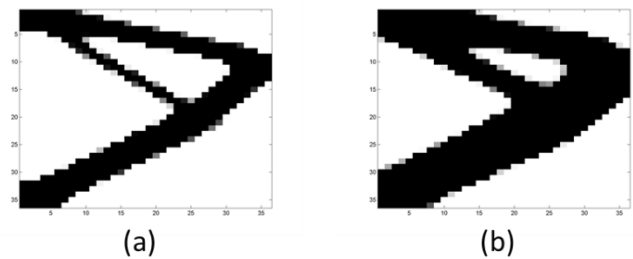

(a)    (b)

Figure 7: Two design solutions with minimal compliance subjected to the same loading condition but with different volume fractions: (a) volume fraction=0.3; (b) volume fraction=0.5.

From the efficiency point of view, the number of the selected design cases should be as few as possible. However, too few cases would yield a dataset that is too small to represent the entire design space for all design cases. As a remedy, data augmentation by altering the existing structures is performed to expand the dataset and add diversity. In particular, density filtering and projection techniques and noise injection are used to expand the dataset. Detailed description is provided in Section 4. We note that a recent work also employs random shape perturbation to expand the dataset (Wang, L. et al. 2020). Once structures are generated, the FEM is used to numerically obtain the ground-truth values. The training of neural networks follows the standard procedure. A second-order optimizer is used in order to achieve high accuracy.

### 3.4 NN-accelerated topology optimization algorithm

Once the dual-model neural network is constructed and trained, it can be integrated into the topology optimization algorithm to speed up the design process. A genetic flow chart of the accelerated TO algorithm is shown in Figure 8. The neural network is used to perform forward calculation and/or sensitivity analysis depending on the type of TO algorithm used. Since the function evaluation on the neural network is much faster than high-dimensional simulations, a great reduction in computational time is expected.

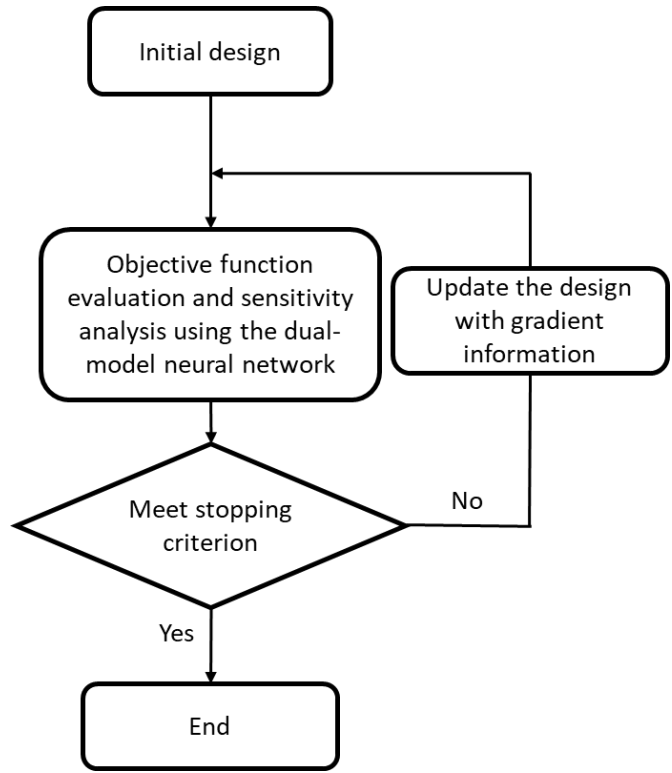

Figure 8: A genetic flow chart of the NN-accelerated topology algorithm.

## 4. Results

### 4.1 Minimal compliance design

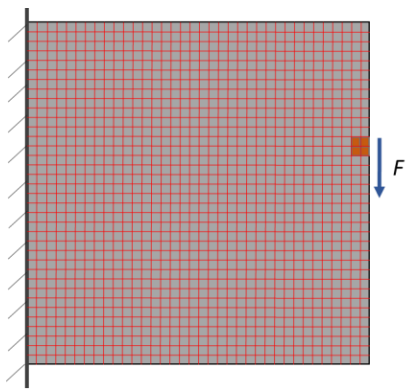

Figure 9: Schematic illustration of the design domain, the boundary conditions and loading condition of the minimum compliance design problem.

The square design domain is discretized into 36 by 36 uniform elements/pixels as shown in Figure 9. A vertically downward uniformly distributed force is applied on a small area occupying 2x2 elements next to the right edge of the domain. For convenience, the force location is indexed by integer numbers from 1 to 18 when it moves from the top four elements to the bottom four elements respectively. For instance, the index of the force location shown in Figure 9 is 8. The solid material used in the design is aluminum with its Young's modulus being $E_0 = 69\ GPa$ and Poisson's ratio being $v_0 = 0.3$. The range of the volume fraction of the solid material considered in this set of design problems is [0.3, 0.6]. First the SIMP combined with the FEM was used to produce a set of benchmark design solutions. Some of them are listed in Table 1. The average number of iterations in each design is 150. As expected, with the increased volume fraction, stiffer structures are produced indicated by smaller compliances.

Table 1: Optimal cantilever designs corresponding to different loading conditions and volume fractions obtained from the SIMP method combined with the FEM.

| Volume fraction | 0.3 | | | 0.4 | | |
|---|---|---|---|---|---|---|
| Loading position | 1 | 9 | 17 | 6 | 10 | 15 |
| Designs | 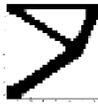 | 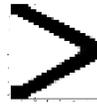 | 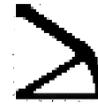 | 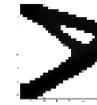 | 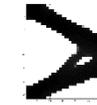 | 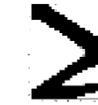 |
| Compliance/$1e^{-3}$J | 2.397 | 1.989 | 2.338 | 1.599 | 1.536 | 1.580 |

| Volume fraction | 0.5 | | | 0.6 | | |
|---|---|---|---|---|---|---|
| Loading position | 4 | 9 | 13 | 5 | 12 | 17 |
| Designs | 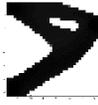 | 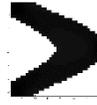 | 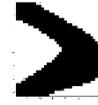 | 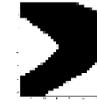 | 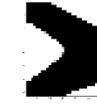 | 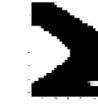 |
| Compliance/$1e^{-3}$J | 1.399 | 1.273 | 1.160 | 0.983 | 0.950 | 1.090 |

To train the dual-model network, 18 design cases are selected in which half of them have the same target volume fraction of 0.35 and odd force indexes ranging from 1 to 17. The other half cases have a target volume fraction of 0.55 and even force indexes ranging from 2 to 18. The SIMP algorithm is run for all 18 cases to produce training structures from different design stages. Five original training datasets are then formed containing structures generated at five different design stages, that is, the early 10%, 30%, 50%, 70% and 90% of the total number of iterations. The corresponding number of structures in each training set is 263, 791, 1319, 1847, 2375 respectively. After training, the neural network is integrated into the SIMP algorithm and used to produce design solutions of testing cases, that is, cases with different loading locations and/or with different volume fractions. A total of 126 cases have been tested. Design solutions of a few cases obtained from the NN-accelerated TO method are listed in Table 2. The corresponding compliances of these designs predicted by neural network are also listed.

Table 2: Optimal cantilever designs obtained from NN-accelerated SIMP methods with NNs constructed based on five original training datasets generated from five different design stages.

| Fraction of design iterations used to generate the training data | 10% | **Target volume fraction** | 0.3 | 0.4 | 0.5 | 0.6 |
|---|---|---|---|---|---|---|
| | | **Loading position** | 2 | 14 | 16 | 6 |
| | | Designs obtained from NN-accelerated TO method | 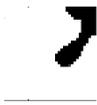 | 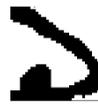 | 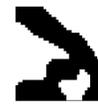 | 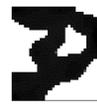 |
| | | Compliance by NN/$1e^{-3}$J | 2.71 | 1.61 | 1.24 | 1.02 |
| | | Prediction error | 100% | 63.32% | 53.52% | 33.57% |
| | | Design error | 3.94e6 | 166.03% | 85.82% | 57.88% |
| | 30% | **Target volume fraction** | 0.35 | 0.4 | 0.55 | 0.6 |
| | | **Loading position** | 5 | 3 | 9 | 18 |
| | | Designs obtained from NN-accelerated TO method | 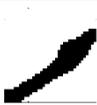 | 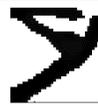 | 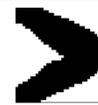 | 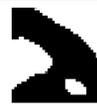 |
| | | Compliance by NN/$1e^{-3}$J | 1.88 | 1.74 | 1.15 | 1.02 |
| | | Prediction error | 98.57% | 53.88% | 11.65% | 13.67% |
| | | Design error | 7329.26% | 112.70% | 0.0% | 1.57% |
| | 50% | **Target volume fraction** | 0.35 | 0.45 | 0.5 | 0.6 |
| | | **Loading position** | 4 | 11 | 2 | 14 |
| | | Designs obtained from NN-accelerated TO method | 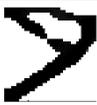 | 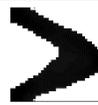 | 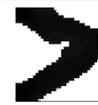 | 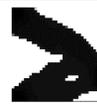 |
| | | Compliance by NN/$1e^{-3}$J | 1.92 | 1.49 | 1.40 | 1.16 |
| | | Prediction error | 17.91% | 5.64% | 100% | 2.69 |
| | | Design error | 27.52% | 12.35% | 8.49e6 | 20.83% |
| | 70% | **Target volume fraction** | 0.3 | 0.4 | 0.45 | 0.55 |
| | | **Loading position** | 18 | 14 | 4 | 14 |
| | | Designs obtained from NN-accelerated TO method | 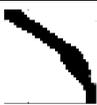 | 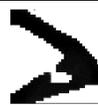 | 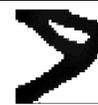 | 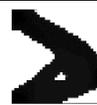 |
| | | Compliance by NN/$1e^{-3}$J | 2.18 | 1.67 | 1.56 | 1256 |
| | | Prediction error | 98.85% | 23.29% | 0.60% | 0.68% |
| | | Design error | 7833.29% | 32.56% | 2.48% | 15.94% |
| | 90% | **Target volume fraction** | 0.35 | 0.45 | 0.55 | 0.6 |
| | | **Loading position** | 2 | 12 | 13 | 18 |

| | | Designs obtained from NN-accelerated TO method | 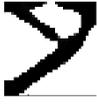 | 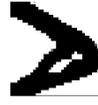 | 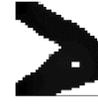 | 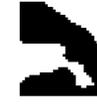 |
|---|---|---|---|---|---|---|
| | | Compliance by NN/$1e^{-3}$J | 2.02 | 1.31 | 1.231 | 1.05 |
| | | Prediction error | 79.48% | 0.04% | 1.04% | 41.50% |
| | | Design error | 387.06% | 0% | 16.23% | 55.14% |

To quantitatively evaluate performances of the dual-model neural network and the accelerated TO method, two types of errors are defined namely the prediction error and the design error. Prediction error measures the prediction accuracy of the dual-model network work and is defined as follows:

$$\text{prediction error} = \frac{|\text{Compliance calculated by FEM} - \text{Compliance calculated by NN}|}{\text{Compliance calculated by FEM}} \quad (18)$$

The prediction error of the compliance corresponding to each design solution is listed in Table 2. Clearly the accuracy improves with the increased number of training data. However this trend stops once the number of training data reaches to a certain value. To provide a clear picture of the relationship between the prediction error and the number of training data, the prediction accuracy of the neural network, averaged over the entire testing set, is plotted against the number of training data collected from TO designs in Figure 10(a). The percentage value in the x-axis indicates the fraction of design iterations used to generate structures. At early design stages, the prediction accuracy increases rapidly with the increased number of structures. But when the fraction of design iterations used increases beyond 30%, the prediction accuracy more or less converges to around 85% and does not improve any further. This trend indicates that structures obtained from later design stages are similar to those generated from early stages, and hence cannot further improve the prediction accuracy of the network.

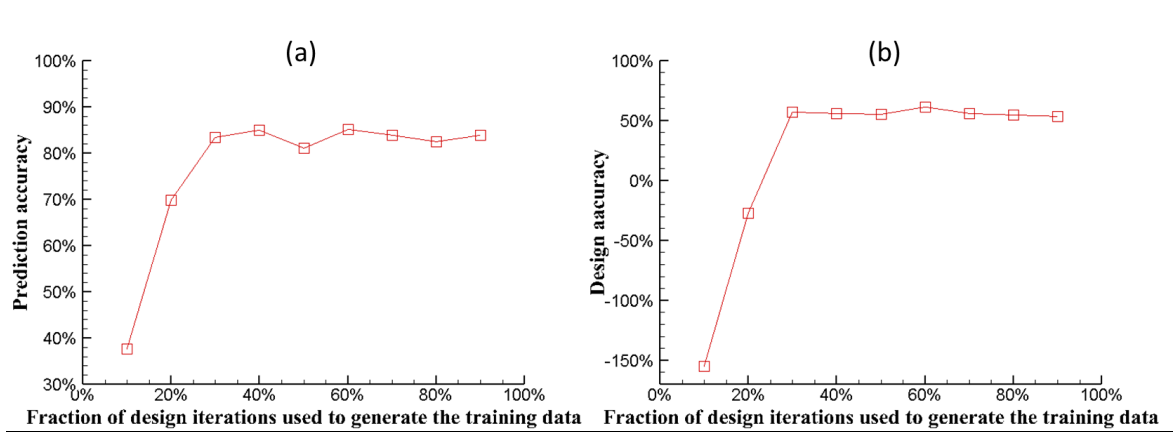

Figure 10: The average prediction accuracy (a) and the average design accuracy (b) of the dual-model neural network versus the fraction of iterations used to generate the original training data.

Design error measures the performance of the NN-accelerated TO method and is defined as:

$$\text{design error} = \frac{\max(\text{Compliacne0} - \text{Compliance1}, 0)}{\text{Compliance1}}, \quad (19)$$

where Compliance0 denotes the compliance of the optimal structure found by the NN-accelerated TO method. This compliance is calculated by the FEM and hence the prediction error is excluded in the design error. Compliance1 denotes the corresponding benchmark solution. In cases when compliance0 is smaller than compliance1, that is, the design solution found by the NN-accelerated TO method is better than that found by the conventional TO method, the design error is set to be zero.

The average design accuracy of the NN-accelerated TO method, which is plotted in Figure 10(b), is greatly affected by the prediction error. The maximum design accuracy is only 60% regardless of the number of training data. As can be seen in Table 2, some designs are barely connected and some with low volume fractions don't even have materials in the region where the external load is imposed. The poor performance of the accelerated TO method is solely due to the low prediction accuracy, which misguides the design process.

To improve the prediction accuracy, more variety of data has to be added to the training set. Given the fact that topologies of structures in different design cases are similar, new structures can be produced simply by altering the existing structures. Density filtering (Bendsøe and Sigmund 2003) and density projection (Wang, F. et al. 2011) are first performed to generate more structures, where the threshold in the projection function is randomly selected from [0.1, 0.9]. Next, noise is added to all structures to further expand the dataset. This is done by adding a random number selected from the range of [-0.2, 0.2] to the density value of each pixel. There are a total of $N/2$ newly generated structures wherer N is the total number of structures generated from 18 TO designs. Half of them are generated from density filtering and projection. The other half are generated from noise injection.

The network is re-trained using augmented datasets and integrated into the SIMP. In Table 3, some of the designs obtained from the accelerated SIMP method are listed together with their compliances. The prediction error and the design error of each case are also listed. Overall, both the prediction and the design accuracies have been largely improved compared with those shown in Table 2. Almost all structures are connected. In particular, when the amount of the training data collected from TO designs increases to a certain level, for instance, 50%, the obtained structures visually look much closer to the benchmark solutions and both the average prediction error and design error are around or smaller than 5%. Similar to what is shown in Table 2, further increasing training data, that is, by adding more structures generated from later design stages, does not further improve both accuracies.

Table 3: Optimal cantilever designs obtained from NN-accelerated SIMP methods with NNs constructed based on augmented training datasets.

|  | 10% | Target volume fraction | 0.40 | 0.45 | 0.55 | 0.6 |
|---|---|---|---|---|---|---|
|  |  | Loading position | 1 | 13 | 13 | 17 |

| Fraction of design iterations used to generate the training data | | | | | | |
|---|---|---|---|---|---|---|
| | | Designs | 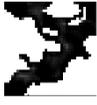 | 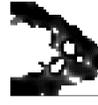 | 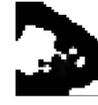 | 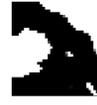 |
| | | Compliance by NN/1e$^{-3}$J | 2.19 | 2.03 | 1.23 | 1.15 |
| | | Prediction error | 11.98% | 2.14% | 3.65% | 11.75% |
| | | Design error | 44.62% | 36.67% | 13.44% | 19.11% |
| | 30% | **Target volume fraction** | 0.3 | 0.4 | 0.5 | 0.6 |
| | | **Loading position** | 13 | 6 | 7 | 17 |
| | | Designs | 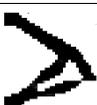 | 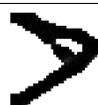 | 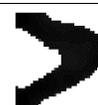 | 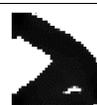 |
| | | Compliance by NN/1e$^{-3}$J | 2.46 | 1.68 | 1.44 | 1.25 |
| | | Prediction error | 14.73% | 8.48% | 10.25% | 1.64% |
| | | Design error | 37.34% | 0.0% | 14.92% | 16.69% |
| | 50% | **Target volume fraction** | 0.35 | 0.4 | 0.55 | 0.6 |
| | | **Loading position** | 1 | 10 | 18 | 15 |
| | | Designs | 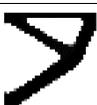 | 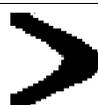 | 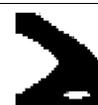 | 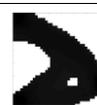 |
| | | Compliance by NN/1e$^{-3}$J | 2.11 | 1.52 | 1.23 | 1.25 |
| | | Prediction error | 0.91% | 6.28% | 2.69% | 5.50% |
| | | Design error | 0.95% | 0.00% | 0.38% | 0.97% |
| | 70% | **Target volume fraction** | 0.30 | 0.35 | 0.40 | 0.55 |
| | | **Loading position** | 12 | 15 | 4 | 12 |
| | | Designs | 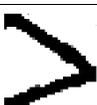 | 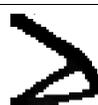 | 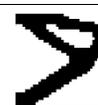 | 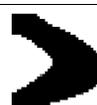 |
| | | Compliance by NN/1e$^{-3}$J | 2.28 | 1.97 | 1.64 | 1.11 |
| | | Prediction error | 2.18% | 2.73% | 1.35% | 8.51% |
| | | Design error | 8.18% | 4.34% | 2.32% | 0.0% |
| | 90% | **Target volume fraction** | 0.30 | 0.45 | 0.5 | 0.55 |
| | | **Loading position** | 4 | 15 | 12 | 9 |
| | | Designs | 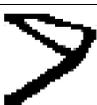 | 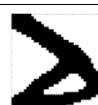 | 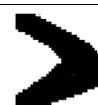 | 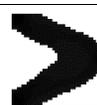 |
| | | Compliance by NN/1e$^{-3}$J | 2.37 | 1.42 | 1.18 | 1.26 |
| | | Prediction error | 1.48% | 1.32% | 4.25 | 5.29 |
| | | Design error | 4.50% | 0.0% | 0.00% | 1.61% |

The prediction and design accuracies averaged over 126 design cases are plotted in Figure 11(a) and (b). With the augmented data, the prediction accuracy can reach to 94% when using only the first 50% of the total structures generated from TO designs. The design accuracy can reach up to 95%, which is a drastic improvement compared with that obtained without augmented data, indicating the effectiveness of data augmentation.

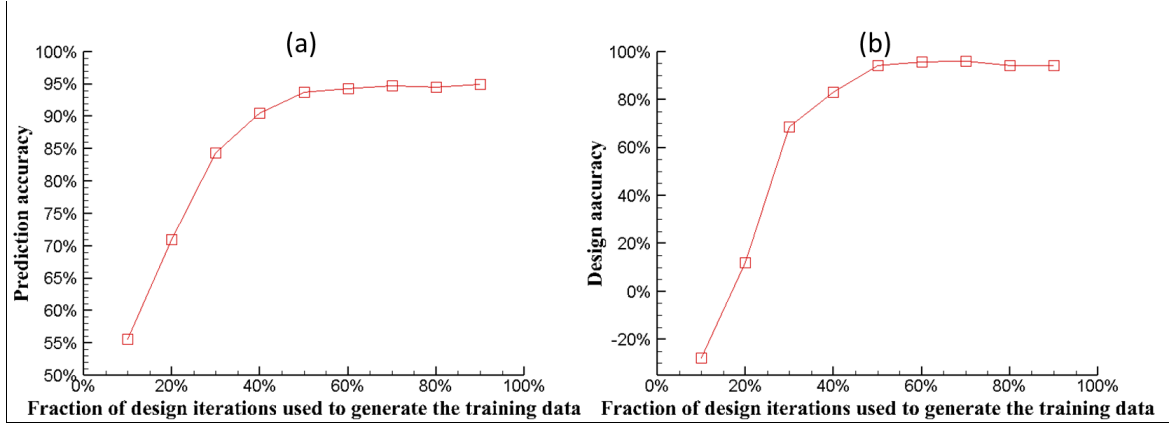

Figure 11: The average prediction accuracy (a) and the average design accuracy (b) of the dual-model neural network trained with augmented datasets versus the fraction of iterations used to generate the original training data.

## 4.2 Metamaterials with negative Poisson's ratios

The design objective in this example is to design a set of microstructures with desired Poisson's ratios in the range of [-0.60, 0.0) and volume fractions in the range of [0.3, 0.65]. The base material chosen for microstructures is aluminum with Young's modulus of $E_0 = 69\ GPa$ and Poisson's ratio of $v_0 = 0.3$. A square design domain is used which is discretized into 64*64 uniform elements. To simplify the optimization problem, symmetric designs are sought and hence only a quarter of the domain is designed. Using the SIMP method combined with the finite element analysis, a set of benchmark designs with various Poisson's ratios and volume fractions are produced. The average number of iterations for each design is around 200. In Table 4, some microstructures and their corresponding Poisson's ratios are listed. The mean relative error of Poisson's ratios of all designs is 2.88% as compared with target values.

Table 4: Metamaterial designs with different Poisson's ratios and volume fractions obtained from the SIMP method combined with the FEM.

| Designs | | | | | | | | |
|---|---|---|---|---|---|---|---|---|
| Volume fraction | 0.3 | | | | 0.4 | | | |
| Desired Poisson's ratio | -0.20 | -0.25 | -0.4 | -0.6 | -0.20 | -0.25 | -0.4 | -0.6 |
| Obtained Poisson's ratio | -0.1952 | -0.2489 | -0.3846 | -0.5971 | -0.1980 | -0.2487 | -0.3880 | -0.5924 |
| Relative error | 2.40% | 0.44% | 3.85% | 0.48% | 1.00% | 0.52% | 3.00% | 1.27% |

| Designs | 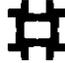 | 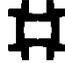 | 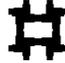 | 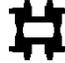 | 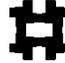 | 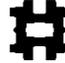 | 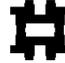 | 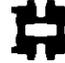 |
|---|---|---|---|---|---|---|---|---|
| Volume fraction | 0.5 | | | | 0.6 | | | |
| Desired Poisson's ratio | -0.20 | -0.25 | -0.4 | -0.6 | -0.20 | -0.25 | -0.4 | -0.6 |
| Obtained Poisson's ratio | -0.1872 | -0.2481 | -0.3990 | -0.5975 | -0.1896 | -0.2408 | -0.3975 | -0.5961 |
| Relative error | 6.40% | 0.76% | 0.25% | 0.42% | 5.20% | 3.68% | 0.63% | 0.65% |

To train the dual-model neural network for the prediction of Poisson's ratio and its sensitivity, 13 design cases are selected to generate training data. Among them, 7 cases have the same prescribed volume fraction of 0.35 but different target Poisson's ratios in the range of [-0.6, 0.0] with an increment of 0.1. The other 6 cases have a target volume fraction of 0.6 and target Poisson's ratios in the range of [-0.55, 0.05] with an increment of 0.1. The SIMP is run for all 13 cases to produce structures from different design stages. Five original trained data sets are then formed containing structures generated at five different design stages, that is, the early 10%, 30%, 50%, 70%,and 90% of the total number of iterations. The corresponding number of structures in each training set is 265, 817, 1353, 1891 and 2434 respectively. After training, these neural networks are integrated into the SIMP to replace FEM analysis. The resulting algorithm is then used to design various microstructures with volume fractions and target Poisson's ratios different from those used in training. A total of 120 design cases have been tested. Some design results and their corresponding Poisson's ratios are listed in Table 5.

Table 5: Metamaterial designs obtained from the NN-accelerated SIMP method with NNs constructed based on five original training datasets generated from five different design stages.

| Fraction of design iterations used to generate the training data | 10% | Volume fraction | 0.325 | 0.380 | 0.525 | 0.590 |
|---|---|---|---|---|---|---|
| | | targeted Poisson's ratio | -0.38 | -0.475 | -0.535 | -0.25 |
| | | Obtained designs | 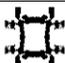 | 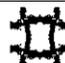 | 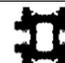 | 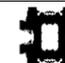 |
| | | Neural network predicted Poisson's ratio | -0.382 | -0.486 | -0.546 | -0.258 |
| | | FEM validated Poisson's ratio | -0.240 | -0.374 | -0.386 | -0.203 |
| | | Prediction error | 59.32% | 30.65% | 41.40% | 27.09% |
| | | Design error | 36.92% | 21.16% | 27.86% | 18.80% |
| | 30% | Volume fraction | 0.325 | 0.430 | 0.525 | 0.620 |
| | | targeted Poisson's ratio | -0.3 | -0.38 | -0.475 | -0.535 |
| | | Obtained designs | 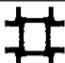 | 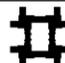 | 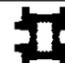 | 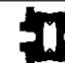 |
| | | Neural network predicted Poisson's ratio | -0.303 | -0.367 | -0.467 | -0.528 |
| | | | -0.267 | -0.324 | -0.423 | -0.418 |

|  |  | Prediction error | 13.5% | 13.51% | 10.31% | 26.35% |
|  |  | Design error | 11.06% | 14.87% | 10.90% | 21.88%% |
|  | 50% | **Volume fraction** | 0.325 | 0.430 | 0.525 | 0.590 |
|  |  | **targeted Poisson's ratio** | -0.535 | -0.38 | -0.125 | -0.475 |
|  |  | Obtained designs | 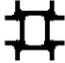 | 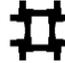 | 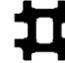 | 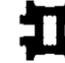 |
|  |  | Neural network predicted Poisson's ratio | -0.511 | -0.366 | -0.116 | -0.461 |
|  |  | FEM validated Poisson's ratio | -0.466 | -0.325 | -0.108 | -0.417 |
|  |  | Prediction error | 9.72% | 12.54% | 7.48% | 10.73% |
|  |  | Design error | 12.92% | 14.44% | 13.37% | 12.27% |
|  | 70% | **Volume fraction** | 0.325 | 0.380 | 0.460 | 0.620 |
|  |  | **targeted Poisson's ratio** | -0.38 | -0.475 | -0.125 | -0.3 |
|  |  | Obtained designs | 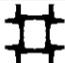 | 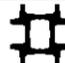 | 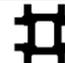 | 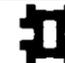 |
|  |  | Neural network predicted Poisson's ratio | -0.367 | -0.476 | -0.121 | -0.296 |
|  |  | FEM validated Poisson's ratio | -0.353 | -0.435 | -0.111 | -0.285 |
|  |  | Prediction error | 4.25% | 9.38% | 9.02% | 3.86% |
|  |  | Design error | 7.12% | 8.39% | 11.47% | 5.04% |
|  | 90% | **Volume fraction** | 0.325 | 0.380 | 0.430 | 0.590 |
|  |  | **targeted Poisson's ratio** | -0.125 | -0.535 | -0.38 | -0.475 |
|  |  | Obtained designs | 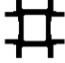 | 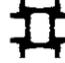 | 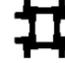 | 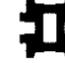 |
|  |  | Neural network predicted Poisson's ratio | -0.113 | -0.535 | -0.364 | -0.470 |
|  |  | FEM validated Poisson's ratio | -0.116 | -0.471 | -0.342 | -0.414 |
|  |  | Prediction error | 2.99% | 13.57% | 6.36% | 13.71% |
|  |  | Design error | 7.21% | 11.87% | 10.03% | 12.90% |

Similar to the previous example, the quantitative evaluation of the design quality are conducted using two measures: the prediction error and the design error defined previously in Equations (18) and (19) except that "compliance" is replaced by Poisson's ratio in this design example. Both errors are listed in in Table 5. In addition, the prediction accuracy of the dual-model neural network method averaged over the 120 design cases is plotted against the fraction of design iterations used to generate the training data in Figure 12(a). A similar trend can be observed in which the prediction accuracy increases rapidly when the fraction of design iterations increase from the first 10% to the first 50%. It is then saturated to around 90% regardless of the increased number of training data.

The average design accuracy, which is plotted in Figure 12(b), also shows the same trend. The highest design accuracy is around 89%, which can be achieved using only the first 50% of the total design iterations to generate original training data.

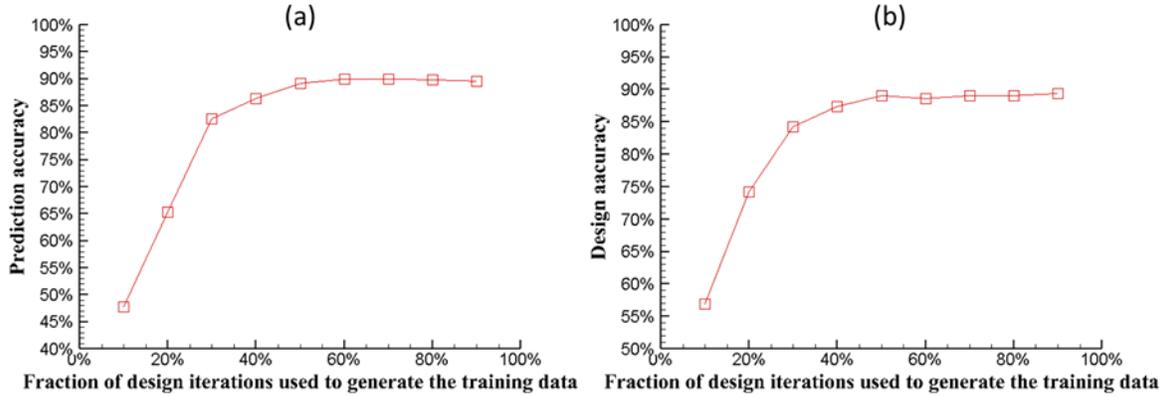

Figure 12: The average prediction accuracy (a) and the average design accuracy (b) of the dual-model NN versus the fraction of iterations used to generate the original training data in the design of metamaterials with negative Poisson's ratios.

To improve the prediction accuracy and design accuracy, data augmentation methods used in the previous example are used to generate more structures. A total of $N/2$ new structures are generated where N is the total number of structures generated from 13 TO designs. Half of the new structures are generated from density filtering and projection. The other half are generated from noise injection. The neural network is retrained with the expanded dataset and integrated into the SIMP method to perform the design of metamaterials with negative Poisson's ratios. Some of the obtained metamaterials are shown in Table 6.

Table 6: Metamaterial designs obtained from the NN-accelerated SIMP method with NNs constructed based on augmented datasets.

|  | 10% | Volume fraction | 0.38 | 0.41 | 0.46 | 0.525 |
|---|---|---|---|---|---|---|
|  |  | targeted Poisson's ratio | -0.125 | -0.38 | -0.3 | -0.36 |
|  |  | Obtained designs |  |  |  |  |
|  |  | Neural network predicted Poisson's ratio | -0.116 | 0.367 | -0.296 | -0.373 |
|  |  | FEM validated Poisson's ratio | -0.153 | -0.267 | -0.171 | -0.214 |
|  |  | Prediction error | 23.66% | 37.78% | 73.36% | 73.94% |
|  |  | Design error | 22.02% | 29.84% | 43.05% | 40.51% |
|  | 30% | Volume fraction | 0.325 | 0.430 | 0.525 | 0.590 |
|  |  | targeted Poisson's ratio | -0.32 | -0.23 | -0.3 | -0.36 |

| Fraction of design iterations used to generate the training data | | Obtained designs | 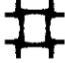 | 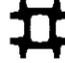 | 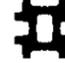 | 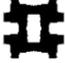 |
|---|---|---|---|---|---|---|
| | | Neural network predicted Poisson's ratio | -0.320 | -0.233 | -0.287 | -0.362 |
| | | FEM validated Poisson's ratio | -0.288 | -0.207 | -0.275 | -0.330 |
| | | Prediction error | 10.98% | 12.23% | 4.29% | 9.62% |
| | | Design error | 9.99% | 9.86% | 8.30% | 8.35% |
| | 50% | **Volume fraction** | 0.38 | 0.43 | 0.48 | 0.59 |
| | | **targeted Poisson's ratio** | -0.17 | -0.125 | -0.425 | -0.3 |
| | | Obtained designs | 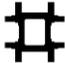 | 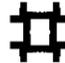 | 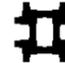 | 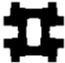 |
| | | Neural network predicted Poisson's ratio | -0.166 | -0.113 | -0.426 | -0.294 |
| | | FEM validated Poisson's ratio | -0.161 | -0.122 | -0.400 | -0.282 |
| | | Prediction error | 3.54% | 7.61% | 6.48% | 4.01% |
| | | Design error | 5.42% | 2.17% | 5.84% | 5.88% |
| | 70% | **Volume fraction** | 0.325 | 0.43 | 0.525 | 0.62 |
| | | **targeted Poisson's ratio** | -0.51 | -0.51 | -0.36 | -0.23 |
| | | Obtained designs | 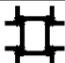 | 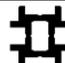 | 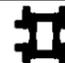 | 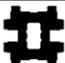 |
| | | Neural network predicted Poisson's ratio | -0.502 | -0.504 | -0.357 | -0.225 |
| | | FEM validated Poisson's ratio | -0.489 | -0.470 | -0.345 | -0.216 |
| | | Prediction error | 2.63% | 5.95% | 3.66% | 4.25% |
| | | Design error | 4.02% | 6.78% | 4.23% | 6.15% |
| | 90% | **Volume fraction** | 0.325 | 0.380 | 0.480 | 0.620 |
| | | **targeted Poisson's ratio** | -0.36 | -0.32 | -0.26 | -0.38 |
| | | Obtained designs | 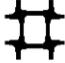 | 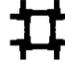 | 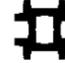 | 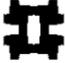 |
| | | Neural network predicted Poisson's ratio | -0.365 | -0.314 | -0.262 | -0.375 |
| | | FEM validated Poisson's ratio | -0.347 | -0.315 | -0.252 | -0.356 |
| | | Prediction error | 5.15% | 4.38% | 3.95% | 5.28% |
| | | Design error | 3.53% | 5.57% | 3.05% | 6.22% |

With the augmented data, the performance of the NN-accelerated SIMP method is much improved. Both the prediction accuracy and the design accuracy averaged over all 120 testing cases increase as shown in Figure 13(a) and (b). Over 94% average prediction accuracy and design accuracy can be achieved by using training data generated from the first 50% of the total iterations. But further increasing the training data does not improve the prediction/design accuracy.

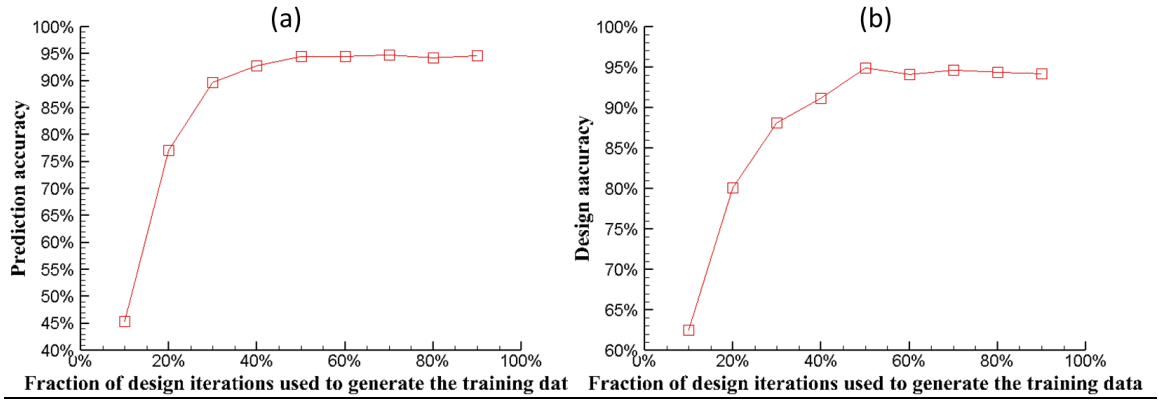

Figure 13: The average prediction accuracy (a) and the average design accuracy (b) of the dual-model NN trained with augmented datasets versus the fraction of iterations used to generate the original training data in the design of metamaterials with negative Poisson's ratios.

## 4.3 Efficiency of the NN-accelerated TO method

One main advantage of the NN-based surrogate model is its efficiency. For the two design examples presented in this paper, the computational costs associated with the forward prediction and sensitivity analysis conducted using dual-model neural networks are listed in Table 7. As a comparison, computational times associated with FEM analysis are also listed. All calculations are conducted on a computer with Intel(R) Xeon(R) CPU E5-2687W v2 (3.40GHz). The computational time for sensitivity analysis using the FEM is similar to that of the forward calculation because the adjoint method is employed. For dual-model neural networks, sensitivity analysis is performed simultaneously with the forward calculation and needs twice amount of the time of the forward calculation. Despite of that, the efficiency of the dual-model neural network is much higher than that of FEA in both forward prediction and sensitivity analysis as indicated in Table 7. For a problem size of 64x64, the efficiency gain in forward calculation is 137 times and in sensitivity analysis is 74 times. It is anticipated that for problems with larger sizes, for example, 3D problems, several orders of magnitude improvement in efficiency can be easily achieved. With NN-surrogate models, TO methods, particularly those require large numbers of iterations, can be greatly accelerated.

Table 7: Computational costs of the dual-model neural network and the FEM for forward and sensitivity calculations.

| Design problem | Problem size | Forward calculation time per structure | | | Sensitivity analysis time per structure | | |
|---|---|---|---|---|---|---|---|
| | | NN (s) | FEM (s) | FEM/NN | NN (s) | FEM (s) | FEM/NN |
| Cantilever design | 36*36 | 0.0039 | 0.21 | 54 | 0.0088 | 0.21 | 24 |

| Metamaterial design | 64*64 | 0.0035 | 0.48 | 137 | 0.0066 | 0.49 | 74 |

Regarding the training time, the total numbers of training data used to achieve a design accuracy of 95% are 1929 in cantilever design and 2030 in metamaterial design respectively. Among them, 610 and 677 are generated by data augmentation methods. The rest is generated by applying the FEM-based SIMP method to a few selected cases. There are a total of 18 selected cases in cantilever design and 13 cases in metamaterial design. These numbers are significantly less than those needed in training a network to map from an intermediate design to the final design.

## 5. Summary and future work

In this work, dual-model neural networks are proposed and used to accelerate gradient-based TO methods for structure/material designs. These neural networks serve as efficient surrogate models to replace high-dimensional numerical simulations employed in conventional TO methods for forward and sensitivity calculations. Two benchmark design examples are successfully demonstrated. In these examples, dual-model neural networks are constructed to predict structure compliance and Poisson's ratio and their sensitivities with respect to the density of each element. Compared with FE analyses, a nearly two orders of magnitude improvement in efficiency has been achieved for a problem size of 60x60. Therefore with the dual-model neural networks, the repetitive forward and sensitivity analyses will no long be the bottleneck that hinders the efficiency of TO methods.

Network training usually requires a significant amount of computational resource. To reduce the amount of training data, effective data generation methods suitable for TO designs are studied and proposed. As a result, only around 2000 training data are required in order to achieve a design accuracy of 95% in each design example. The resulting fast SIMP methods can perform minimum compliance design and metamaterial design with prescribed volume fraction and loading condition or Poisson's ratio in a wide range.

The design examples tested in this work are two-dimensional linear problems. The proposed method, that is, using dual-model neural networks as the surrogate models, should be applicable to three dimensional and/or nonlinear problems. In fact, the efficiency gained would be much higher in 3D and nonlinear problems. Although the SIMP method is used in this work to demonstrate the acceleration performance of NN-based surrogate models, these surrogate models can be easily incorporated into non-gradient TO methods, for example, genetic algorithm based TO methods. Since the number of iterations in non-gradient based TO methods is typically much higher than that of the gradient-based methods, the acceleration achieved would be much more significant.

# Replication of results

The training data used in this work can be found at https://github.com/hkust-ye/cqian_dual-model_neural_network/tree/master.

**Funding information:** This work is supported by the Hong Kong Research Grants under Competitive Earmarked Research Grant No. 16212318.

**Compliance with ethical standards**

**Conflict of interest:**   The authors declare that they have no conflict of interest.